\documentclass[final]{cvpr}

\usepackage{times}
\usepackage{epsfig}
\usepackage{graphicx}
\usepackage{dsfont}
\usepackage{amsmath}
\usepackage{amsfonts}
\usepackage{amssymb}
\usepackage{algpseudocode}
\usepackage{dblfloatfix}
\usepackage{subfig}
\usepackage[titlenumbered,ruled,linesnumbered]{algorithm2e}
\SetAlFnt{\small}

\usepackage{amsmath,amsfonts,bm}

\def\eqref#1{equation~\ref{#1}}

\def\1{\bm{1}}

\DeclareMathAlphabet{\mathsfit}{\encodingdefault}{\sfdefault}{m}{sl}
\SetMathAlphabet{\mathsfit}{bold}{\encodingdefault}{\sfdefault}{bx}{n}

\DeclareMathOperator*{\argmax}{arg\,max}
\DeclareMathOperator*{\argmin}{arg\,min}

\usepackage[pagebackref=true,breaklinks=true,colorlinks,bookmarks=false]{hyperref}

\newcommand{\footlabel}[2]{%
    \addtocounter{footnote}{1}%
    \footnotetext[\thefootnote]{%
        \addtocounter{footnote}{-1}%
        \refstepcounter{footnote}\label{#1}%
        #2%
    }%
    $^{\ref{#1}}$%
}

\newcommand{\footref}[1]{%
    $^{\ref{#1}}$%
}

\def\cvprPaperID{****} 
\def\confYear{EarthVision 2021}

\begin{document}

\title{Training Domain-invariant Object Detector Faster\\with Feature Replay and Slow Learner}

\author{Chaehyeon Lee\textsuperscript{1}\thanks{Work done as intern at SI Analytics.} \hspace*{1cm} Junghoon Seo\textsuperscript{2} \hspace*{1cm} Heechul Jung\textsuperscript{1}\thanks{Corresponding author.}\\
\textsuperscript{1}Department of Artificial Intelligence, Kyungpook National University, Daegu, Korea \\ \textsuperscript{2} SI Analytics, Co., Ltd., Daejeon, Korea\\
{\tt\small \textsuperscript{1}\{123456ccdd, heechul\}@knu.ac.kr \textsuperscript{2}jhseo@si-analytics.ai}}
\clearpage\maketitle
\thispagestyle{empty}
\pagestyle{empty}

\begin{abstract}
In deep learning-based object detection on remote sensing domain, nuisance factors, which affect observed variables while not affecting predictor variables, often matters because they cause domain changes.
Previously, nuisance disentangled feature transformation (NDFT) was proposed to build domain-invariant feature extractor with with knowledge of nuisance factors.
However, NDFT requires enormous time in a training phase, so it has been impractical.
In this paper, we introduce our proposed method, A-NDFT, which is an improvement to NDFT. A-NDFT utilizes two acceleration techniques, feature replay and slow learner. Consequently, on a large-scale UAVDT benchmark, it is shown that our framework can reduce the training time of NDFT from  31 hours to 3 hours while still maintaining the performance.
The code will be made publicly available online\footnote{\url{https://github.com/2-Chae/A-NDFT}}.
\end{abstract}

\section{Introduction}
\begin{figure}[t!]
\begin{tabular}{c}
\includegraphics[width = \columnwidth]{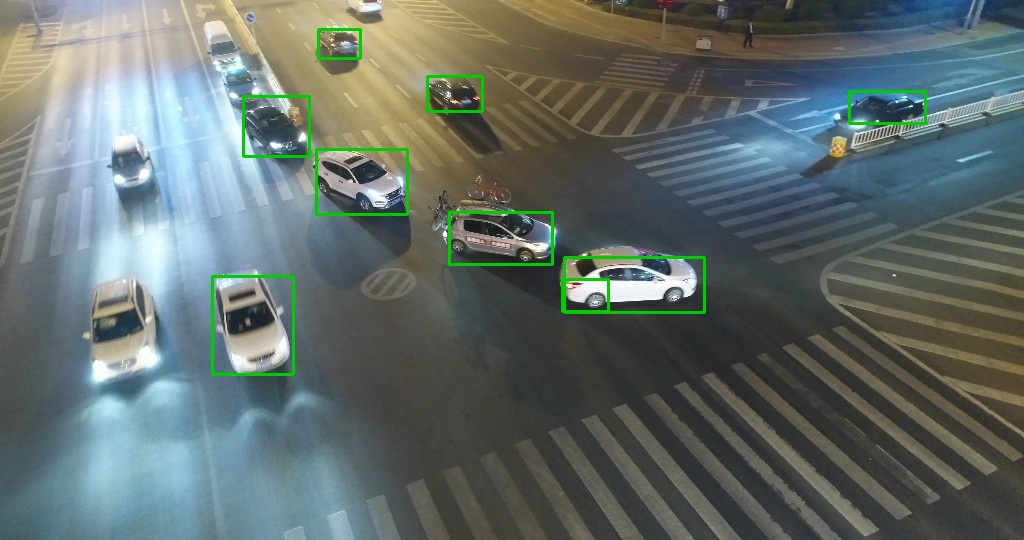} \\
(a) Baseline\\[6pt] 
\includegraphics[width = \columnwidth]{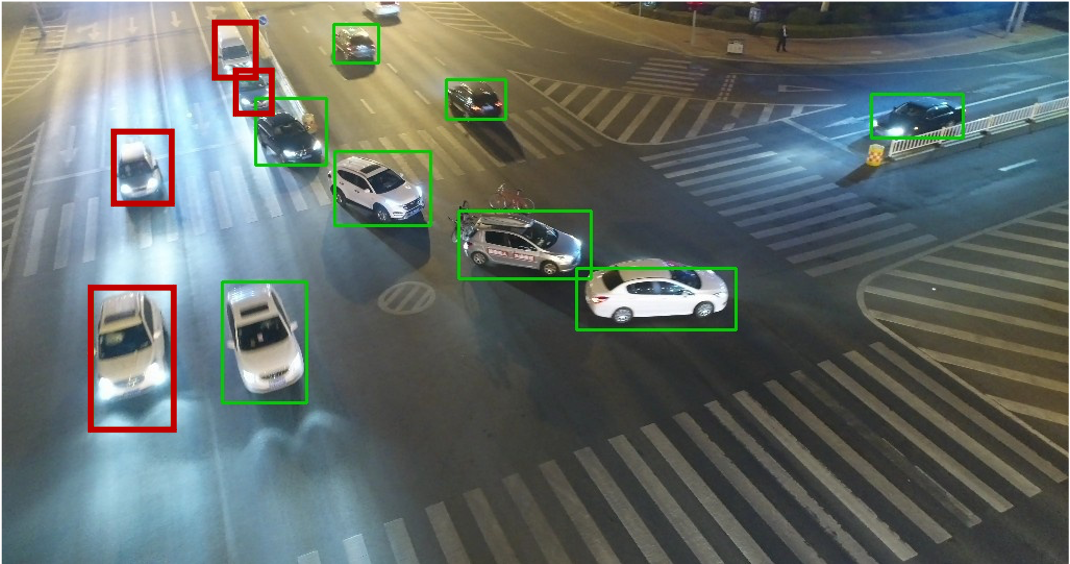}\\
(b) A-NDFT \\[6pt] 
\end{tabular}
\vspace{-1.0em}
\caption{Detection examples and comparison with Faster R-CNN baseline and the corresponding A-NDFT. Red boxes denote new correct detections where the baseline failed to detect, but A-NDFT did.}
\label{visdrone_vis}
\vspace{-1em}
\end{figure}

Nuisance factors matter everywhere in deep learning-based object detection in remote sensing field.
In imagery obtained from an unmanned aerial vehicle (UAV), shooting height, shooting angle, and shooting time significantly influence the appearance of scene and objects \cite{du2018unmanned,zhu2020vision}.
In imagery obtained from satellites, the type of satellite, ground sample distance, azimuth/altitude angle of the satellite, azimuth/altitude angle of the sun, and various environmental factors work in the same way \cite{christie2018functional,weir2019spacenet}.
These nuisance variables are marginally independent from target variables, but affect observation variables \cite{NIPS2017_8cb22bdd,tenenbaum1997separating,li2014learning}.
Therefore, the learned representation in the deep object detector can be dependent on nuisance variables.
It induces over-fitting biased towards frequently appearing nuisance variables.

In the perspective of domain shift, the issue of nuisance factors in object detection has been addressed from various domain adaptation approaches.
\cite{khodabandeh2019robust} addressed the domain adaptation problem from the perspective of noise-resistant robust learning.
\cite{dwibedi2017cut,dvornik2018modeling,tripathi2019learning} leveraged image composition between seen object instances and unseen scenes.
\cite{xu2020cross} proposed a novel graphical framework on category-level domain alignment.
\cite{lee2019me} adopted multiple experts to capture variations of the object appearance better.
Each of these methodologies has revealed its strengths, but most of them are not suitable to deal with the nuisance factors appearing in the remote sensing field described above.
This is because these methods cannot handle massive numbers of fine-grained domains.

Instead, in this paper, we focus on the target method, nuisance disentangled feature transformation (NDFT) \cite{wu2019delving}.
Motivated from privacy-preserving visual recognition \cite{wu2018towards} and disentangled feature learning \cite{Liu_2018_CVPR}, NDFT makes use of an adversarial learning framework to build a domain-robust object detector.
Unlike many of the previously mentioned methodologies, it is suitable for dealing with the massive number of nuisance factors.
Moreover, to the best of our knowledge, NDFT's original literature is the pioneering work to show the effectiveness of fine-grained domain-invariant learning in UAV images.
\cite{wu2019delving} reported that in the UAVDT benchmark accompanying the Faster R-CNN model, the introduction of NDFT had increased mAP by over 2\%.
It results from using only meta-data that can be obtained for free without any additional labeling work.
For this reason, NDFT is expressed as 'free lunch' in their paper.

Contrary to this attraction, in this paper, we point out that the domain robustness obtained with NDFT is not actually a free lunch.
In other words, the use of NDFT requires a lot of training time.
We analyze the causes of these phenomena and suggest alternative strategies to train NDFT models.
Specifically, our contributions in this paper are as follows:
\begin{itemize}
  \item We pointed out the problem of slow training of NDFT and analyzed the causes of it.
  \item Based on the problem analysis, we present an alternative learning methodology for NDFT. One is the feature replay, and another is the slow learner.
  \item The experiment results in the UAVDT benchmark show that our proposed A-NDFT can significantly reduce the training time while maintaining the performance of the existing NDFT.
\end{itemize}

\section{Related Works}
\subsection{Domain Shift of Remote Sensing}
Domain shift problem is ubiquitous in remote sensing applications.
Several works have reported that machine learning models on remote sensing suffered domain shift caused by various nuisance factors, e.g., viewpoint geometry, atmospheric effect, temporal variability, and sensor properties \cite{czaja2018adversarial}.
In \cite{weir2019spacenet}, it was discussed that most of the existing satellite datasets were collected with only the "at nadir" data and thus showed inferior performance in off-nadir observation data.
\cite{kiefer2021leveraging,wu2019delving} observed the performance drop of object detector according to the view angle, which is trained on VisDrone dataset \cite{zhu2020vision} and UAVDT dataset \cite{du2018unmanned}.
\cite{tasar2020standardgan} observed the phenomenon that the model performance deteriorated due to the change in color distribution even when sensing different regions with the same satellite.

Under awareness of the problem, a great deal of research has dealt with domain adaptation problem in remote sensing \cite{song2019domain,lu2019multisource,deng2019large,koga2020method,tasar2020daugnet,tasar2020standardgan}.
Most of the existing work has dealt with domain adaptation for classification or semantic segmentation.
Moreover, these studies have assumed a multi-source domain adaptation situation in which the source domain and the target domain are entirely separated.
On the contrary, our study is based on the NDFT model, so we aim to construct a robust object detector in fine-grained domains.

\subsection{Adversarial Learning for Domain Invariance}
There are quite a number of studies using adversarial frameworks to impart domain invariance to deep learning models \cite{pmlr-v37-ganin15,xie2017controllable,NEURIPS2018_717d8b3d,ijcai2020-271,jaiswal2020invariant,yuan2020calibrated,pmlr-v119-wang20h}.
\cite{pmlr-v37-ganin15} introduced domain-adversarial neural network (DANN), an adversarial framework between domain predictor and feature extractor.
In DANN, the feature extractor is trained to decrease the task loss in the source domain and increase the loss of the domain predictor.
In \cite{xie2017controllable}, adversarial learning for training the feature extractor was performed in the form of maximizing the entropy of the domain predictor's output.
In addition, studies on domain invariance through adversarial learning have been dealt with in various aspects such as neural network architecture \cite{jaiswal2020invariant}, extension to multiple domains \cite{NEURIPS2018_717d8b3d}, stabilization techniques of training \cite{ijcai2020-271,yuan2020calibrated}, and methods of dealing with continuous domains \cite{pmlr-v119-wang20h}.

Similarly, our study explores the NDFT model that acquires domain invariance through adversarial learning.
Therefore, it can be seen that this study also deals with this category of topics.
Nevertheless, at the same time, our work only focuses on convergence speed and acceleration techniques of the NDFT model.
Therefore, our research has a point that is differentiated from the viewpoint of the existing researches.

\section{Problem Formulation and Motivations}
\begin{algorithm*}[ht]
\SetKwInOut{Input}{Input}
\SetKwInOut{Output}{Output}
\SetKwInOut{Required}{Required}
\SetKwInOut{Defined}{Defined}
\Input{$f_T(\cdot;\theta_T): \mathcal{X} \mapsto \tilde{\mathcal{X}}$, a backbone network \newline
$f_O(\cdot;\theta_O): \tilde{\mathcal{X}} \mapsto \mathcal{P}$, a module for object detection task branch \newline
$\{f_{N,i}(\cdot;\theta_{N,i})\}_{i=1}^{k}: \tilde{\mathcal{X}} \mapsto \mathcal{Y_N}$, an ordered set of $k$ modules for the nuisance prediction branches \newline
$D$, $M$-size training dataset consisting of each triplet $(X, Y_O, Y_N)$
}
\Required{$\{ \gamma_i \}_{i=1}^{k}$, an ordered set of weight coefficients \newline
$T$, a number of training iterations \newline
$\alpha$, a threshold value for nuisance prediction performance \newline
$\psi$, a hyper-parameter indicating restart cycle of nuisance predictor \newline
$\eta_U$ and $\eta_N$, two scalar values for learning rate
}
\Defined{$\theta_U \gets \theta_T \cup \theta_O$ and $\theta_{N.U} \gets \theta_{N,1} \cup \cdots \cup \theta_{N,k}$
}
\Output{$f_O(f_T(\cdot;\theta_T);\theta_O): \mathcal{X} \mapsto \mathcal{P}$, domain-invariant object detector}
\vspace{-0.2em}
\caption{NDFT: Nuisance Disentangled Feature Transform}\label{algo:ndft}
\vspace{0.5em}
\For{$t$ $\mathrm{ \textbf{in} }$ $\mathrm{range(}T\mathrm{)}$}{
Sample a mini-batch of $n$ data $\{(X^j, Y_O^j, Y_N^j)\}_{j=1}^n$ from $D$\;
$\theta_U \gets \theta_U - \eta_U \nabla_{\theta_U} \frac{1}{n}\displaystyle\sum_{j=1}^n \Big[L_O(f_O(f_T(X^j)), Y_O^j)+\displaystyle\sum_{i=1}^{k}\gamma_i L_{ne}(f_{N,i}(f_T(X^j)))\Big]$\; \label{line:ndtf-3}
\While{$\min_{1 \le i \le k} \Big[ \frac{1}{n}\displaystyle\sum_{j=1}^n [\mathds{1}(\argmax(f_{N,i}(f_T(X^j))) = Y_{N,i}^j)] \Big] \le \alpha$ \label{line:ndtf-4}}{
$\theta_{N,U} \gets \theta_{N,U} - \eta_N \nabla_{\theta_{N,U}}\frac{1}{n}\displaystyle\sum_{j=1}^n\displaystyle\sum_{i=1}^k{L_N(f_{N,i}(f_T(X^j)),Y_{N,i}^j)}$\; \label{line:ndtf-5}
} \label{line:ndtf-6}
\If{$t \mathrm{  } \% \mathrm{  } \psi = 0$ \label{line:ndtf-7}}{
Re-initialize $\theta_{N,U}$\; \label{line:ndtf-8}
}\label{line:ndtf-9}
 }
\vspace{-0.2em}
\end{algorithm*}
\vspace{-0.5em}

\subsection{Domain-invariant Feature for Object Detection}
Let $\mathcal{X}$ be the domain of the input image and $\tilde{\mathcal{X}}$ be the space of the intermediate features.
The domain to represent localization and classification results of each possible region obtained by the detection task is denoted as $\mathcal{P}$.
The backbone network $f_T(\cdot;\theta_T): \mathcal{X} \mapsto \tilde{\mathcal{X}}$ takes the input image $X \in \mathcal{X}$ and produces an intermediate feature $f_T(X;\theta_T)$.
The detection task network $f_O(\cdot;\theta_O): \tilde{\mathcal{X}} \mapsto \mathcal{P}$ receives this feature and calculates classification and localization for all possible region candidates.
When the two networks are combined (i.e. $f_O(f_T(\cdot;\theta_T);\theta_O):\mathcal{X} \mapsto \mathcal{P}$), it can be interpreted as a generic object detector.
By abuse of notation, we do not consider the difference between one-stage \cite{liu2016ssd,zhou2019objects} and two-stage detectors \cite{ren2015faster} for simplicity.
In addition, the model parameter is omitted when it is obvious enough that it is not necessary to specify it, e.g., $f_T(\cdot;\theta_T) = f_T(\cdot)$.
The loss function for training this object detector is denoted as $L_O(\cdot,\cdot):\mathcal{P} \times \mathcal{Y_O} \mapsto \mathbb{R}$, and $L_O$ is often defined as the weighted sum of the localization loss and the classification loss.

In our setup, we have $D$, $M$-size dataset as the training data.
Every single instance of D is composed of triplet $(X, Y_O, Y_N)$.
$X \in \mathcal{X}$, $Y_O \in \mathcal{Y_O}$, and $Y_N \in \mathcal{Y_N}$ are information about image data, object detection labeling, and nuisance factors, respectively.
For $Y_N$ to be a nuisance factor, $Y_O$ and $Y_N$ must be marginally independent.
This means that the generation process for the observation of $X$ is dependent on $Y_O$ and $Y_N$, but $Y_O$ and $Y_N$ are independently generated.
For example, in object detection on an outdoor UAV image, the location and scale of the detection target are usually independent of whether it is day or night at the time of shooting.
However, the appearance of the image may depend on the object in the image and the photographed time.

Typically, the object detector $f_O(f_T(\cdot))$ is trained to map from the image $X$ to the object label $Y_O$.
It can over-fit because of $X$'s dependence on $Y_N$.
The variation of $X$ due to the change in $Y_N$ is expressed as $X^\prime$, and $f_O(f_T(X))$ may be different from $f_O(f_T(X^\prime))$.
If the vast majority of image data included in the training dataset were taken during the day, this model might perform poorly on images taken at night.
This intuition motivates the feature extractor's domain-invariant properties.
In other words, we would like to make the backbone network $f_T(\cdot)$ to have the domain-invariant property $f_T(X) = f_T(X^\prime)$.

\subsection{Nuisance Disentangled Feature Transform}
\label{sec:ndft}

For building a domain-invariant object detector, \cite{wu2019delving} formulated fine-grained cross-domain object detection as an adversarial training framework \cite{goodfellow2014generative}.
Motivated by privacy-preserving visual recognition \cite{wu2018towards}, they proposed three-party game among where three players, a backbone network $f_T(X;\theta_T)$, a detection task network $f_O(\cdot;\theta_O)$, and a nuisance predictor $f_N(\cdot:\theta_N):\tilde{\mathcal{X}} \mapsto \mathcal{Y_N}$.

The nuisance predictor $f_N$ is usually a softmax classifier that predicts $Y_N$ by receiving features from the backbone network.
Therefore, the nuisance predictor $f_N$ is trained to minimize the loss $L_N$, which is the surrogate of the nuisance prediction performance, e.g., multi-class cross-entropy loss.
Conversely, the backbone network $f_T$ and the detection task network $f_O$ are trained to maximize the nuisance prediction loss $L_N$ while minimizing the object detection task loss $L_O$.
When considering multiple nuisance prediction tasks, the training procedure of the three modules is formulated with an alternative optimization of the following two objectives:
\begin{align}\label{obj:NDFT_multiple}
\displaystyle&\argmin_{\theta_O, \theta_T} L_O(f_O(f_{T}(X)), Y_O)-\displaystyle\sum_{i=1}^{k}\gamma_i L_N (f_{N,i}(f_{T}(X)), Y_{N, i}), \nonumber \\
\displaystyle&\argmin_{\theta_{N,1},\cdots,\theta_{N,k}} L_N(f_{N,i}(f_{T}(X)), Y_{N,i})
\vspace{-3em}
\end{align}
where $k$ denotes the number of nuisance factors, and the $i$-th elements related to them are denoted as $(\cdot)_{N,i}$. $\gamma_i \in \mathbb{R}_{\ge 0}$ is weight coefficient for balancing $L_O$ and $L_N$.

To avoid convergence problems in GAN training \cite{pmlr-v119-farnia20a}, the authors of \cite{wu2019delving} designed a sophisticated training strategy.
This novel training strategy includes the use of negative entropy loss \cite{liu2018multi}, performance monitoring of nuisance predictors, and re-starting tricks \cite{wu2018towards}.
They named the $f_T(\cdot;\theta_T)$ learned by the training loop, including this modification as the Nuisance Disentangled Feature Transform (NDFT).
Algorithm \ref{algo:ndft} depicts the main training loop of NDFT, and the three essential details described in each line are:
\begin{enumerate}
  \item Line \ref{line:ndtf-3}: Instead of using gradient reversal trick \cite{pmlr-v37-ganin15} (i.e., using $-L_N$), the negative entropy of the softmax vector, $L_{ne}$, are adopted as the adversarial loss. This option was also introduced in other works \cite{liu2018multi,yuan2020calibrated}.
  \item Line \ref{line:ndtf-4}-\ref{line:ndtf-6}: To prevent each $f_{N,i}$ from becoming too weak, the training performances of all $k$ nuisance prediction tasks are monitored. In a single alternative optimization, $\theta_{N,U}$ is updated until the prediction accuracy of all nuisance tasks is greater than $\alpha$.
  \item Line \ref{line:ndtf-7}-\ref{line:ndtf-9}: To help prevent falling into bad local minima, we reinitialize $\theta_{N,U}$ every $\psi$ iterations. It is motivated from \cite{wu2018towards}.
\end{enumerate}
The original work of NDFT reported that by using (almost free) meta-data, the performance of the standard Faster R-CNN \cite{ren2015faster} in the UAVDT benchmark dataset \cite{du2018unmanned} could increase mAP above 2\%.
For more details on NDFT, refer to its original work \cite{wu2019delving}.

\subsection{Slow Convergence of NDFT}
Although several UAV object detection benchmarks revealed NDFT's effectiveness, we point out its serious issue for practice usage.
That is, NDFT requires enormous training time.
In our early experiments, we tried to run the author's official code of NDFT\footlabel{ndft-url}{\url{https://github.com/VITA-Group/UAV-NDFT}}, and we found that the training time of NDFT on UAVDT dataset was above 30 hours.
In contrast, we also found that training their baseline, a standard Faster R-CNN model, required only about three hours.
This training time difference indicates that NDFT is not truly 'free lunch' in the aspect of computation and time cost.

We point to several causes of NDFT's slow training speed.
First, the nuisance prediction parameters $\theta_{N,U}$ is updated with the backbone network parameters $\theta_{U}$ fixed.
Thus, repeated feed-forward operations through the backbone network (i.e., $f_T(X^j)$) in Line \ref{line:ndtf-5} result in computational inefficiency.
Second, in the object detection network, it usually needs a lot of training iteration to fine-tune specific predictors with the parameters of the backbone network fixed \cite{wang2020few}.
Considering the periodic parameter initialization heuristics in Line \ref{line:ndtf-7}-\ref{line:ndtf-9} together, the strict performance monitoring policy (Line \ref{line:ndtf-4}) can significantly reduce the overall training speed.
As mentioned in Section \ref{sec:ndft}, they control the over-fitting and under-fitting of the nuisance predictor.
Therefore, it is not easy to achieve both this adjustment and fast training by simply controlling hyper-parameters $\alpha$ and $\psi$.
We may need another strategy to achieve rapid training while adjusting them appropriately.

\section{Our Approach for Acceleration of NDFT}
\begin{figure*}[t!]
\begin{center}
\includegraphics[width=0.95\linewidth]{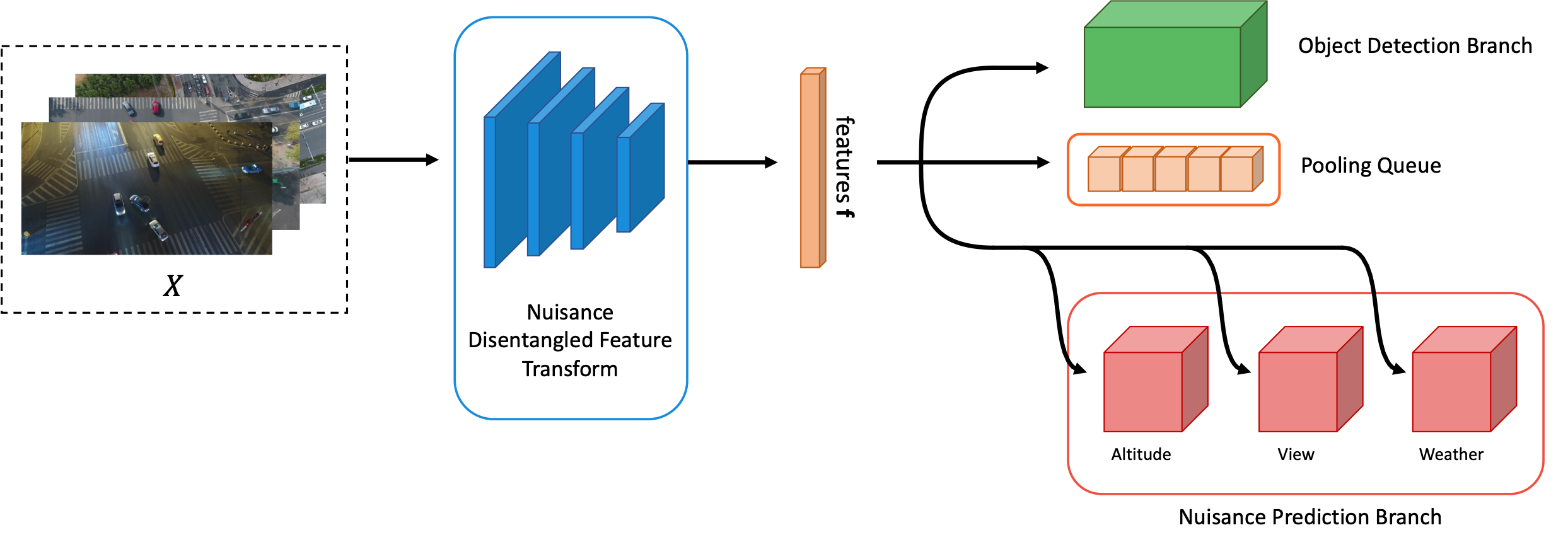}
\end{center}
\caption{Our proposed framework (A-NDFT). We have added a pooling queue used for feature replay to the original NDFT framework. The encoded feature will be used for object detection and nuisance prediction, just as NDFT. Not only that, we store the feature in the pooling queue to avoid an additional feed-forward process of the backbone network.}
\label{fig:structure} 
\end{figure*}

\begin{figure*}[t!]
\begin{center}
\includegraphics[width=0.9\linewidth]{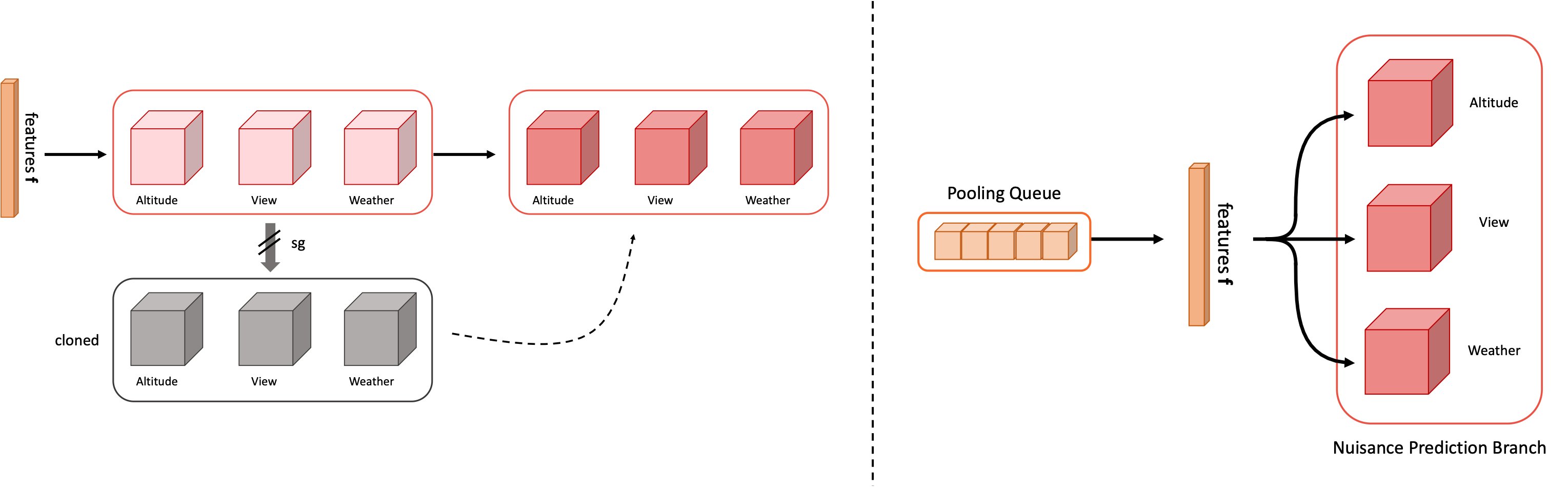}
\end{center}
\caption{Training procedure of nuisance prediction branches. We adopt exponential moving average on parameters to avoid forgetting past information while accepting current information. Here, \textit{sg} means stop gradient (Left: slow runner). When training nuisance branches for each iteration, we use features extracted and stored in the pooling queue when we learned object detection branches from the previous step (Right: feature replay).}
\label{fig:nuisance} 
\end{figure*}


\begin{algorithm*}[ht]
\SetKwInOut{Input}{Input}
\SetKwInOut{Output}{Output}
\SetKwInOut{Required}{Required}
\SetKwInOut{Defined}{Defined}
\Input{Same with Algorithm \ref{algo:ndft}
}
\Required{$\{ \gamma_i \}_{i=1}^{k}$, an ordered set of weight coefficients \newline
$T$, a number of training iterations \newline
$Q_s$, a $s$-size FIFO queue for restoring intermediate features \newline
$\beta$, the degree of weighting decrease for slow learning \newline
$\phi$, a hyper-parameter to indicate training cycle using $Q_s$ \newline
$\eta_U$ and $\eta_N$, two scalar values for learning rate
}
\Defined{$\theta_U \gets \theta_T \cup \theta_O$ and $\theta_{N,U} \gets \theta_{N,1} \cup \cdots \cup \theta_{N,k}$
}
\Output{$f_O(f_T(\cdot;\theta_T);\theta_O): \mathcal{X} \mapsto \mathcal{P}$, domain-invariant object detector}
\vspace{-0.2em}
\caption{A-NDFT: Our Modified Version of Nuisance Disentangled Feature Transform}\label{algo:a-ndft}
\vspace{0.5em}
\For{$t_o$ $\mathrm{ \textbf{in} }$ $\mathrm{range(}T\mathrm{)}$}{
Sample a mini-batch of $n$ data $\{(X^j, Y_O^j, Y_N^j)\}_{j=1}^n$ from $D$\;
$\mathbf{F} \gets \{f_T(X^j)\}_{j=1}^n$ and $Q_s\mathrm{.enqueue(}\mathbf{F}\mathrm{)}$\;
$\theta_U \gets \theta_U - \eta_U \nabla_{\theta_U} \frac{1}{n}\displaystyle\sum_{j=1}^n \Big[L_O(f_O(F_j), Y_O^j)+\displaystyle\sum_{i=1}^{k}\gamma_i L_{ne}(f_{N,i}(F_j))\Big]$ where $F_j = f_T(X^j)$\;
$\theta_{N,U} \gets \beta \theta_{N,U} + (1 - \beta) \Big\{ \theta_{N,U} - \eta_N \nabla_{\theta_{N,U}}\frac{1}{n}\displaystyle\sum_{j=1}^n\displaystyle\sum_{i=1}^k{L_N(f_{N,i}(F_j),Y_{N,i}^j)} \Big\}$\;\label{a-ndft-5}
\If{$t \% \phi = 0$\label{a-ndft-6}}{
\For{$t_i$ $\mathrm{ \textbf{in} }$ $\mathrm{range(floor(}Q_s\mathrm{.num\_of\_items}/n\mathrm{))}$}{
Sample a mini-batch of $n$ features $\{F^j\}_{j=1}^n$ from $Q_s$\ where each $F^j \in \tilde{\mathcal{X}}$\;
$\theta_{N.U} \gets \theta_{N,U} - \eta_N \nabla_{\theta_{N,U}}\frac{1}{n}\displaystyle\sum_{j=1}^n\displaystyle\sum_{i=1}^k{L_N(f_{N,i}(F^j),Y_{N,i}^j)}$\;
}
}
 \label{a-ndft-11}}
\vspace{-0.2em}
\end{algorithm*}
\vspace{-0.5em}

\subsection{Feature Replay}
NDFT \cite{wu2019delving} extracts features two times during each training iteration when training (1) NDFT module and object detection module and (2) nuisance prediction modules. It uses different mini-batches of examples, respectively. However, we point out that the NDFT module's feature extraction process is computationally expensive and time-consuming. So here we introduce feature replay, which avoids this redundancy and guarantees faster learning.

We implement feature replay as a pooling queue containing features from the past steps. During each training iteration, features are extracted from the feature extractor, and they are used as an input of the object detection branch and the nuisance prediction branches, respectively. Before moving on to the next step, we store these features in the pooling queue (Figure \ref{fig:structure}). Note that we store features as many as a batch size at once; the computational gain is also dependent on the batch size. When training the nuisance prediction branches, we do not extract features but use the top element of the queue, thereby cutting down on training time. However, in this way, the nuisance prediction branches might lose information about past inputs as they only learn mini-batches at that point.  We train the nuisance prediction branches with all the elements in the queue for every $k$ iterations to learn without losing (Figure \ref{fig:nuisance}, Right).

Our feature replay can be seen as similar to 'image history' in SimGAN \cite{shrivastava2017learning} or 'experience replay' in reinforcement learning \cite{mnih2015human} and continual learning \cite{rolnick2018experience}.
These are intended to avoid catastrophic forgetting \cite{robins1995catastrophic} of neural networks and stabilize the training process.
However, unlike other studies, our feature replay focuses on the redundancy of computation and stores data points in feature space $\tilde{\mathcal{X}}$, not in data space $ \mathcal{X}$.

\subsection{Slow Learners}

We notice that the nuisance prediction branches' performances keep fluctuating when training because, for every iteration, we train the branches with the top element of the queue of which the nuisance attributes may not be evenly distributed. As a result, they continue to forget past information while accepting the current ones. We want them to learn slowly to make sure they are sufficiently trained, so we adopt exponential moving average (EMA) of parameters. We use EMA operation on nuisance predictors' parameters to keep the average representation and accept new data simultaneously, which slowly trains the model.

As in the left of Figure \ref{fig:nuisance}, we have model parameters before an update. We clone and annotate them as $\theta$. These parameters $\theta$ are not updated during training; for that reason, we denote stop-gradient (sg) operation \cite{grill2020bootstrap} in that figure. Let the current model parameters be $\xi$ which have equivalent to $\theta$ at this moment. Now we compute the parameters of the next step as:
\begin{equation}
\label{eq:2}
    \xi \leftarrow \beta\xi + (1 - \beta)\theta
\end{equation}
where $\beta$ is a decay rate between 0 and 1. 

Although this update method has almost the same context as decreasing the learning rate in finding the center point between the current and historical parameters, we can still reduce the effort of finding the appropriate hyper-parameters.
Moreover, our slow learner is closely related to the EMA-GAN \cite{yazici2019unusual}.
However, unlike EMA-GAN, which aims to increase the stability and performance of the generative model, we show that slow learning of the adversarial predictors is also effective in fine-grained domain adaptation.

\vspace{-0.5em}
\begin{table*}[ht!]
\small
\vspace{-0.5em}
\centering 
\resizebox{1.7\columnwidth}{!}{
\begin{tabular}{c|c|c|c|c|c|c|c|c||c||c} 
\hline
{} & Baseline \cite{ren2015faster} & A & V & W & A+V & A+W & V+W & A+V+W & A+V+W (NDFT) & A+V+W \cite{wu2019delving}\\ 
\hline
& \multicolumn{8}{c}{Flying Altitude}\\
\hline
Low & 54.19    & 55.06 & 52.48 & 55.17 &   46.68  & 55.27 & 55.64 & \textbf{56.91} & 57.46 & 74.84\\ 
\hline
Med & 39.48    & 43.04 & 39.13 & 44.74 & 34.76 & 45.16 & 38.62 & \textbf{46.20} & 46.60 & 56.24 \\
\hline
High & 7.54     & 10.10 & 5.82  & 11.06 &  5.65 & 11.15 & \textbf{11.42} & 8.42 & 10.10 & 20.55 \\ 
\hline
& \multicolumn{8}{c}{Camera View}\\
\hline
Front & 47.10    & 48.03 & 46.58 & 48.07 &  39.24 & 48.40 & 47.51 & \textbf{52.86} &53.26 & 64.88\\ 
\hline
Side & 37.87    & 39.33 & 37.29 & 39.16 &  30.44 & 38.90 & 36.79 & \textbf{39.56} &40.11 & 67.50\\ 
\hline
Bird  & 73.35    & 66.40 & 61.55 & 68.44 & 58.58 & 67.57 & \textbf{76.30} & 72.89 & 86.71 & 28.79\\
\hline
& \multicolumn{8}{c}{Weather Condition} \\
\hline
Day & 45.50    & 47.39 & 41.30 & 47.45 & 38.14 & 48.43 & 45.39 & \textbf{49.24} & 49.40 & 45.91\\
\hline
Night & 49.18    & 49.19 & 43.79 & 50.51 &  41.53 & 48.14 & 49.40 & \textbf{51.13} &53.07 & 64.16 \\
\hline
\hline
Overall & 45.59    & 46.51 & 43.90 & 46.82 &37.90& 47.00 & 45.43 & \textbf{48.12} & 48.37 & 47.91 \\
\hline
\end{tabular}}
\caption{Performance of A-NDFT-Faster-RCNN with multiple attribute disentanglement. Note, A+V+W (NDFT) is the result of reproduced NDFT with our settings and A+V+W \cite{wu2019delving} is the original result from \cite{wu2019delving}.} 
\label{combined-tbd} 
\end{table*}

\begin{figure*}[t!]
\begin{tabular}{*{3}{@{\hspace{5px}}c}}
  \includegraphics[width=56mm]{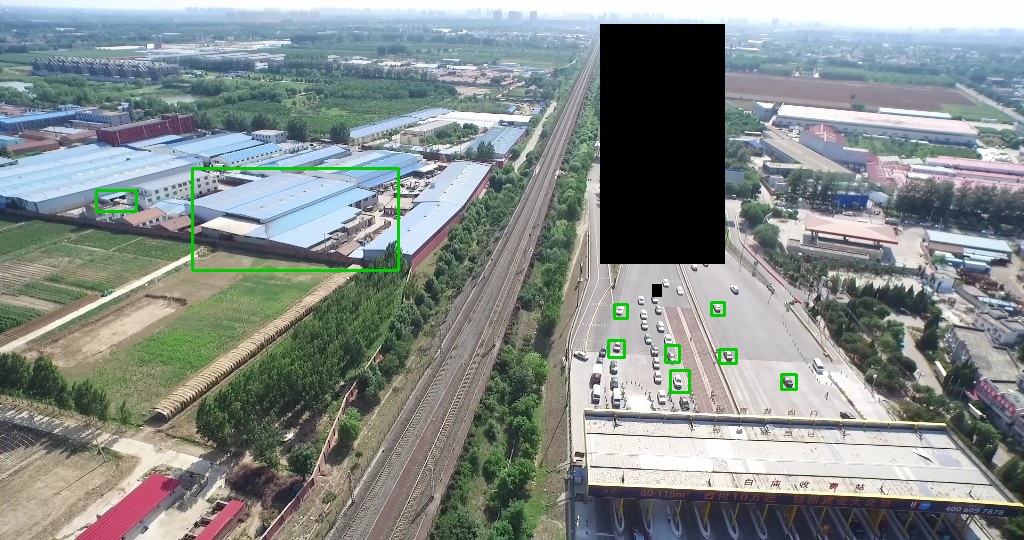} &  
  \includegraphics[width=56mm]{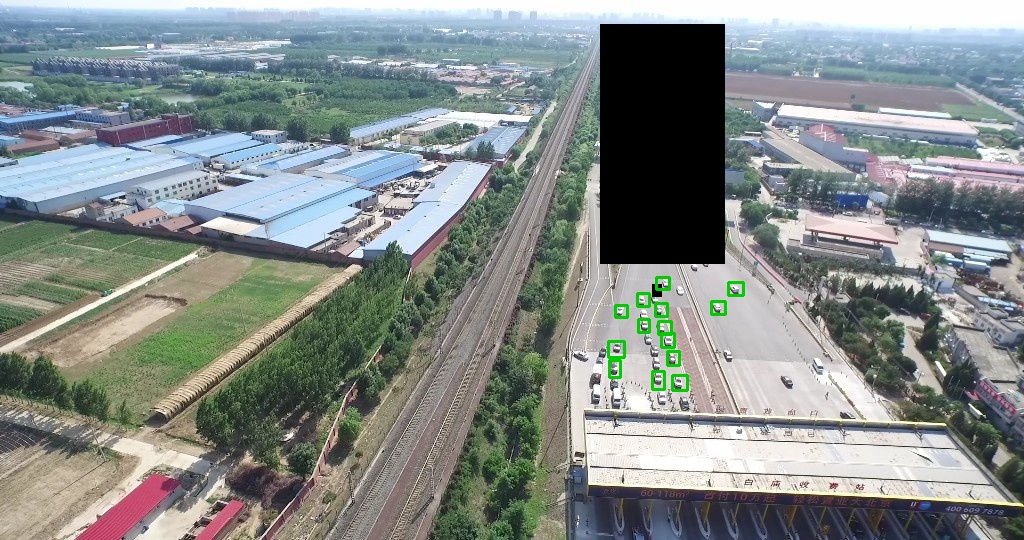} &  \includegraphics[width=56mm]{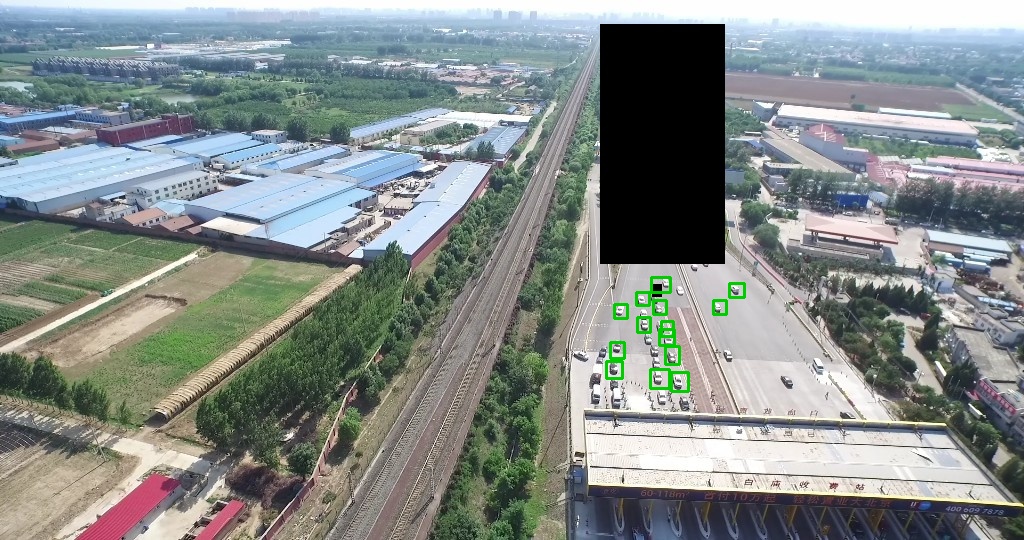} \\
 \includegraphics[width=56mm]{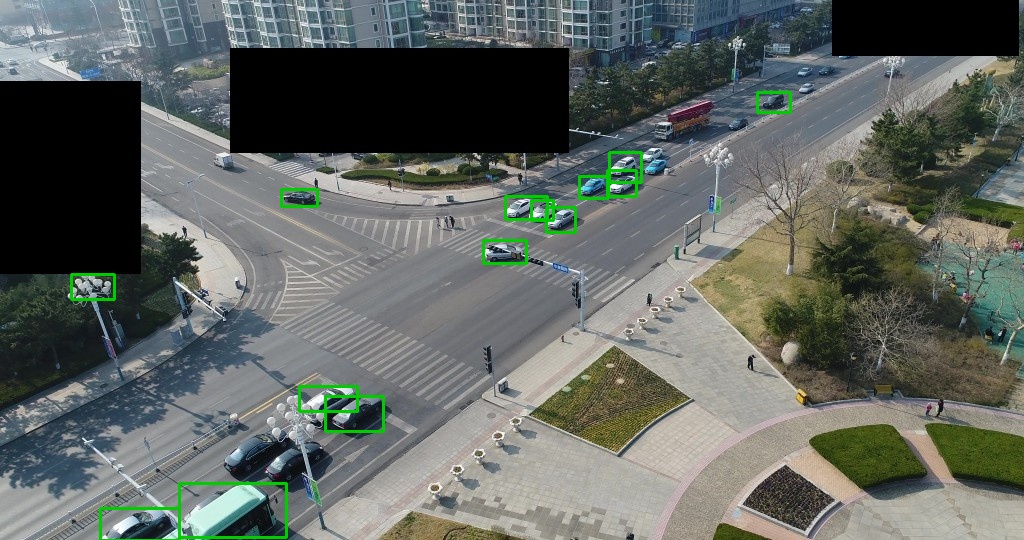} &  
 \includegraphics[width=56mm]{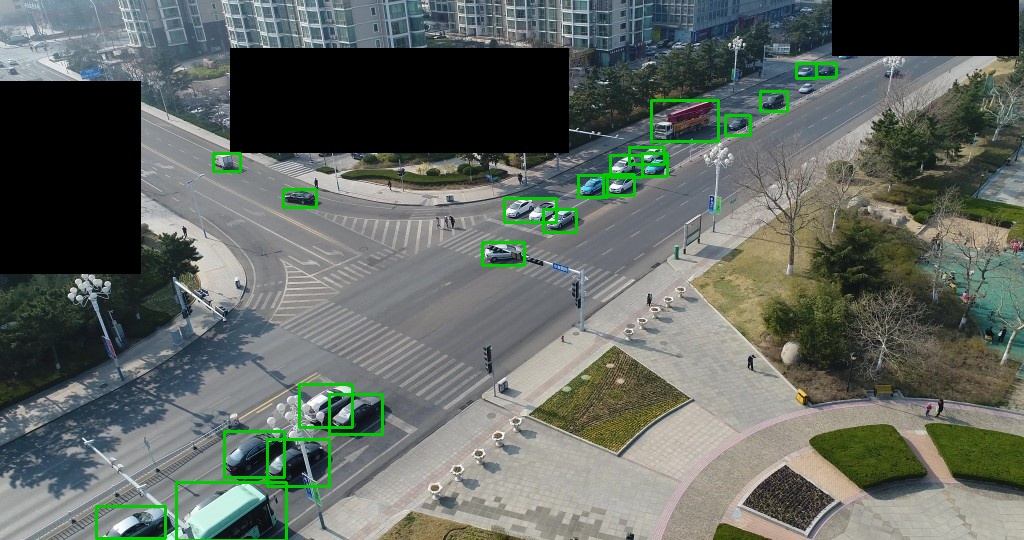} &  \includegraphics[width=56mm]{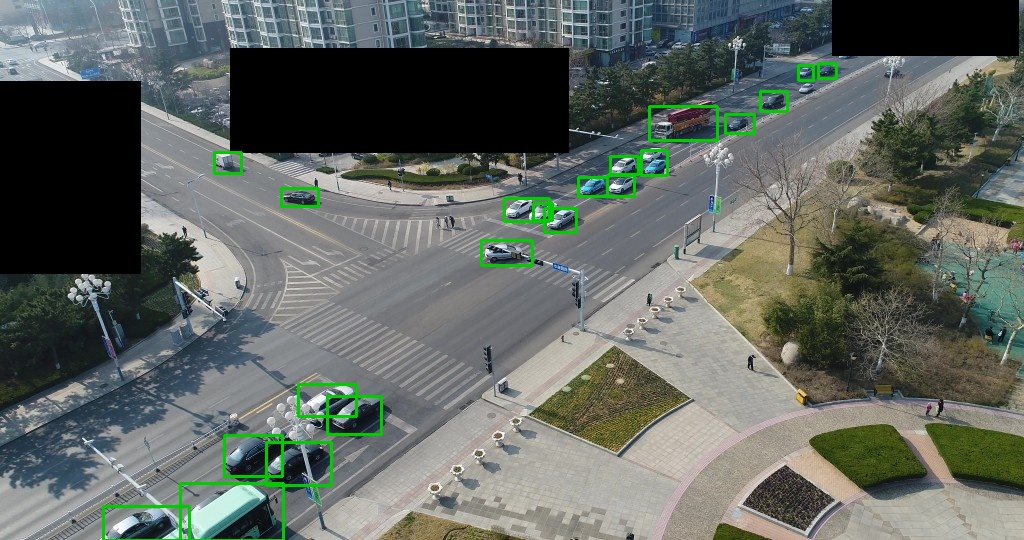} \\
  \includegraphics[width=56mm]{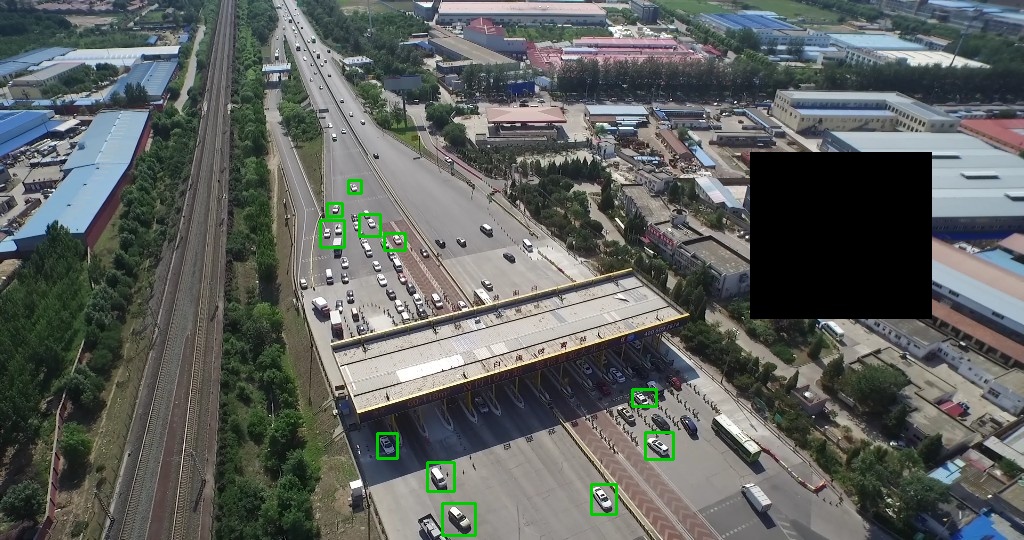} &  
 \includegraphics[width=56mm]{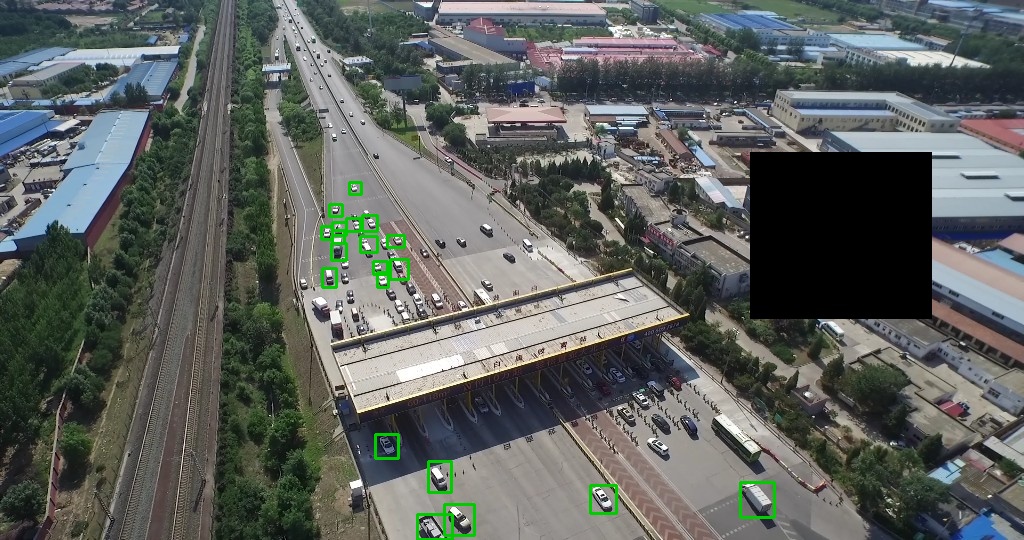} &  \includegraphics[width=56mm]{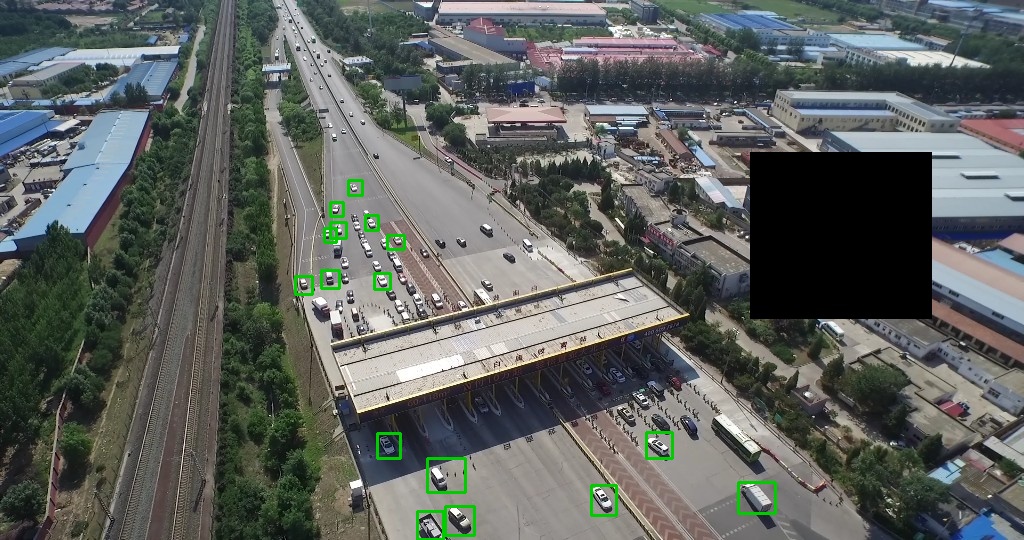} \\
(a) Baseline&(b) NDFT & (c) A-NDFT \\[6pt]
\end{tabular}
\caption{Examples of the proposed A-NDFT that performs object detection on the UAVDT benchmark. Our A-NDFT framework performs better than the baseline and shows comparable performance to NDFT. Best viewed in color at high-resolution (black rectangles in figures are ignored regions provided by UAVDT benchmark itself).}
\label{fig:examples}
\end{figure*}

\subsection{Overall Algorithm}

Here, we present an overall training procedure of our proposed algorithm, A-NDFT. We handle the model training process delicately so that there is no performance difference from the existing technique NDFT but shorten training times. This procedure is summarized in Algorithm \ref{algo:a-ndft} and explained below.

As our A-NDFT is a faster version of the original NDFT, we follow almost all the steps from them. 
Just like NDFT \cite{wu2019delving}, we jointly optimize the backbone model and object detection branch by minimizing the same objective equation as NDFT does.
At this point, we store extracted features in the pooling queue to avoid the redundant feed-forward process of the backbone network. 
After that, NDFT tried to keep monitoring nuisance prediction branches so that their performance does not go down.
With its performance monitoring strategy, we observed the fluctuation of these branches' performance while training.
We assumed that this comes from an imbalance of mini-batch examples and oblivion of past information.
To overcome this, we decided to train nuisance prediction branches with the top element of the queue and apply the slow learner technique, the exponential moving average update on module weights.
By updating weights to average points of current and past weights, modules slowly forget less and learn new information (Algorithm \ref{algo:a-ndft} Line \ref{a-ndft-5}).
Besides, to guarantee the adequate performances of branches, we retrain the modules with whole elements of the pooling queue every $\phi$ iterations (Algorithm \ref{algo:a-ndft} Line \ref{a-ndft-6}-\ref{a-ndft-11}).

\section{Experiments}
\subsection{Experiment Settings} \label{exp_setting}
\paragraph{Dataset: UAVDT \cite{du2018unmanned}} 
The Unmanned Aerial Vehicle Detection and Tracking (UAVDT) is large-scale object detection and tracking benchmark dataset obtained by UAVs. The objects of interest in this benchmark are vehicles annotated over 2,700 with 0.84 million bounding boxes. This benchmark dataset has about 80k frames consisted of 100 video sequences captured from  UAVs in various scenarios.
Moreover, it provides fully-annotated 14 kinds of attributes for each frame.
We use only three categories of them for evaluating our model on the object detection track on UAVDT; flying altitudes, weather conditions, and camera views.  

Flying Altitudes have three levels; low, medium, and high. Weather conditions include daylight, night, and fog. Nevertheless, we did not use fog examples just as NDFT because of its small size. Camera views have three different views, $i.e.$, front-view, side-view, and bird-view. In addition to these three views, there are examples with a front-side view in the dataset. These are frames taken at the border between front and side. The processing of them is not specified in \cite{wu2019delving}, so we decide to merge them into side-view category. By merging frames with front-side view attribute into side-view frames, the number of frames belonging to each attribute can differ from NDFT as a whole, which can raise a discrepancy with NDFT's performance. We will denote these three nuisances as \textbf{A}, \textbf{W} and  \textbf{V}, respectively.

\paragraph{Implementation Details}
We used the same codebase of the NDFT's official repository\footref{ndft-url}, a Faster-RCNN model using ResNet-101 as a backbone model. $\gamma_1$, $\gamma_2$, and $\gamma_3$ denote the coefficients for altitude, view, and weather nuisances, respectively.
To maintain their official setting as unchanged as possible, we initially tried to train the model for five epochs and size-up the batch to 32 and the learning rates $\eta_U$ and $\eta_O$ to 0.08.
However, we only report the NDFT's performance up to three epochs because it takes more than two days to train NDFT for five epochs and NDFT showed a tendency to converge at three epochs. In the case of A-NDFT, we trained the model for five epochs and set the size of the pooling queue as $s=256$.
Nevertheless, even in this setting, A-NDFT ends up 10x faster compared to NDFT. We set hyper-parameters indicating the degree of weighting $\beta$ decrease for slow learning and a training cycle $\phi$ using the entire pooling queue as 0.99 and 325, respectively. Also, we trained A-NDFT and baseline using the same batch size and learning rate as NDFT.

When training the baseline by setting all $\gamma$s as 0, we obtained an AP of 45.59\% (using IoU threshold = 0.7), similar to 45.64\%, which is reported in \cite{wu2019delving}.
Since this model is equivalent to the standard Faster R-CNN model, it is indicated as the baseline.


\subsection{Detection Performance of A-NDFT}
We perform the experiments on UAVDT for object detection. First, we study the impact of just using each nuisance type (\textbf{A}, \textbf{V}, and \textbf{W}) and then combine those into two or three nuisance types. 
As \cite{wu2019delving} reported, applying $\gamma=0.01$ results in the best performance; we also apply all the values of $\gamma_i$ as 0.01 in this experiment. Table \ref{combined-tbd} shows the overall results by incrementally adding adversarial losses to training. \textbf{A}, \textbf{V}, and \textbf{W} are when the branch corresponding to flying altitude, camera view, and weather condition nuisance is used for training. For example, \textbf{A} is when set $\gamma_V, \gamma_W$ to 0 and $\gamma_A$ not to 0. \textbf{A+V+W} represents the experiment when all three nuisance branches are used in training, which means all $\gamma$s are not 0.

When training with only one nuisance module, the performance was the best when using \textbf{W}, and in the case of two branches, the performance was the best when using \textbf{A+W}. Their improvements over the baseline are +1.23\% and +1.41\%, respectively. When all the branches were used, we could see that the performance was the best in all cases. For a detailed comparison, we added the performance of NDFT from \cite{wu2019delving}, and a reproduced NDFT to the rightmost side of the table. The discrepancy with the value in \cite{wu2019delving} could arise from the differences in the data preprocessing as mentioned in \ref{exp_setting}. Precisely, the number of frames belonging to each attribute may not match  \cite{wu2019delving}’s setting, so there are some differences in per-nuisance performances. However, the overall case has a similar result to the performance of  \cite{wu2019delving} because it trains using all data without considering attributes. Consequently, the performance difference between A-NDFT and NDFT is not significant, which means that A-NDFT maintains the performance of NDFT. Figure \ref{fig:examples} shows some visual examples for a qualitative comparison.

\subsection{Convergence Speed: A-NDFT versus NDFT}
\begin{figure}[t!]
\centering
\includegraphics[width=8cm]{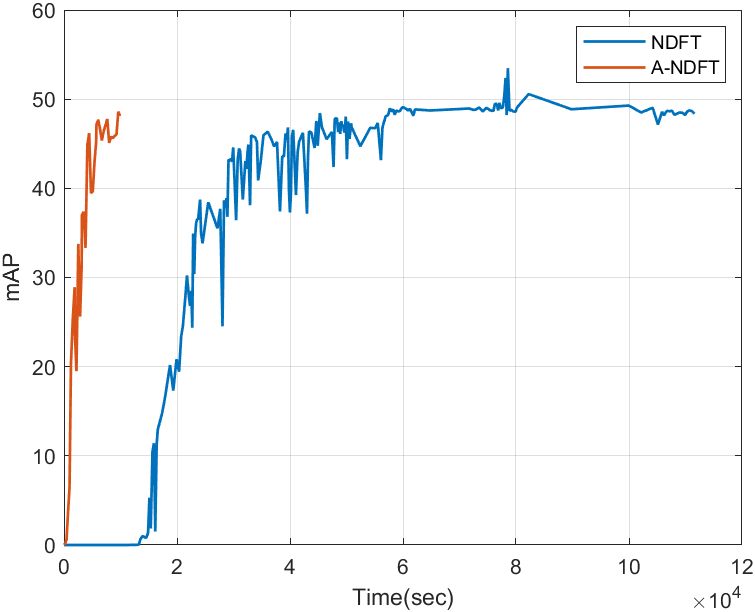}\label{fig1}
\caption{A comparison of mAP over training time for A-NDFT and NDFT. Note that the NDFT was trained for three epochs, whereas our A-NDFT was trained for five epochs.}
\label{fig:convergence} 
\end{figure}

In this section, we give a comparison of convergence speeds of A-NDFT and NDFT. We use the same architecture specified in \ref{exp_setting}. In training NDFT, based on the implementation of \cite{wu2019delving}, we increased the batch size by 10x and reduced the epochs proportionally to make the most of GPU resources and cut down on the learning time. NDFT took about 31 hours to train, while A-NDFT finished in about 3 hours. We visualize their performance over time in Figure \ref{fig:convergence}. By utilizing the feature replay and the slow learner,  A-NDFT significantly reduces training time and speeds up convergence than NDFT.

\section{Conclusion}
In this paper, we present A-NDFT, which uses a feature replay and a slow learner to accelerate the training of NDFT. The proposed method saves ten times as much time as and still maintains performance analogous to NDFT. We conducted vehicle detection on the UAVDT dataset, a large-scale benchmark, and showed that it performs better than baseline, not using nuisance factor predictions, and performs comparably with NDFT. In the future, we intend to carry out further analyses of various remote-sensing datasets such as and VisDrone \cite{zhu2020vision} and use the other backbone architectures.

\section*{Acknowledgements}
This work was supported by the Defense Challengeable Future Technology Program of the Agency for Defense Development, Republic of Korea.

{\small
\bibliographystyle{ieee_fullname}
\bibliography{egbib}

\begin{thebibliography}{10}\itemsep=-1pt

\bibitem{christie2018functional}
Gordon Christie, Neil Fendley, James Wilson, and Ryan Mukherjee.
\newblock Functional map of the world.
\newblock In {\em Proceedings of the IEEE Conference on Computer Vision and
  Pattern Recognition}, pages 6172--6180, 2018.

\bibitem{czaja2018adversarial}
Wojciech Czaja, Neil Fendley, Michael Pekala, Christopher Ratto, and I-Jeng
  Wang.
\newblock Adversarial examples in remote sensing.
\newblock In {\em Proceedings of the 26th ACM SIGSPATIAL International
  Conference on Advances in Geographic Information Systems}, pages 408--411,
  2018.

\bibitem{deng2019large}
Xueqing Deng, Hsiuhan~Lexie Yang, Nikhil Makkar, and Dalton Lunga.
\newblock Large scale unsupervised domain adaptation of segmentation networks
  with adversarial learning.
\newblock In {\em IGARSS 2019-2019 IEEE International Geoscience and Remote
  Sensing Symposium}, pages 4955--4958. IEEE, 2019.

\bibitem{du2018unmanned}
Dawei Du, Yuankai Qi, Hongyang Yu, Yifan Yang, Kaiwen Duan, Guorong Li, Weigang
  Zhang, Qingming Huang, and Qi Tian.
\newblock The unmanned aerial vehicle benchmark: Object detection and tracking.
\newblock In {\em Proceedings of the European Conference on Computer Vision
  (ECCV)}, pages 370--386, 2018.

\bibitem{dvornik2018modeling}
Nikita Dvornik, Julien Mairal, and Cordelia Schmid.
\newblock Modeling visual context is key to augmenting object detection
  datasets.
\newblock In {\em Proceedings of the European Conference on Computer Vision
  (ECCV)}, pages 364--380, 2018.

\bibitem{dwibedi2017cut}
Debidatta Dwibedi, Ishan Misra, and Martial Hebert.
\newblock Cut, paste and learn: Surprisingly easy synthesis for instance
  detection.
\newblock In {\em Proceedings of the IEEE International Conference on Computer
  Vision}, pages 1301--1310, 2017.

\bibitem{pmlr-v119-farnia20a}
Farzan Farnia and Asuman Ozdaglar.
\newblock Do {GAN}s always have {N}ash equilibria?
\newblock In Hal~Daumé III and Aarti Singh, editors, {\em Proceedings of the
  37th International Conference on Machine Learning}, volume 119 of {\em
  Proceedings of Machine Learning Research}, pages 3029--3039. PMLR, 13--18 Jul
  2020.

\bibitem{pmlr-v37-ganin15}
Yaroslav Ganin and Victor Lempitsky.
\newblock Unsupervised domain adaptation by backpropagation.
\newblock In Francis Bach and David Blei, editors, {\em Proceedings of the 32nd
  International Conference on Machine Learning}, volume~37 of {\em Proceedings
  of Machine Learning Research}, pages 1180--1189, Lille, France, 07--09 Jul
  2015. PMLR.

\bibitem{goodfellow2014generative}
Ian~J Goodfellow, Jean Pouget-Abadie, Mehdi Mirza, Bing Xu, David Warde-Farley,
  Sherjil Ozair, Aaron Courville, and Yoshua Bengio.
\newblock Generative adversarial networks.
\newblock {\em arXiv preprint arXiv:1406.2661}, 2014.

\bibitem{grill2020bootstrap}
Jean-Bastien Grill, Florian Strub, Florent Altch{\'e}, Corentin Tallec,
  Pierre~H Richemond, Elena Buchatskaya, Carl Doersch, Bernardo~Avila Pires,
  Zhaohan~Daniel Guo, Mohammad~Gheshlaghi Azar, et~al.
\newblock Bootstrap your own latent: A new approach to self-supervised
  learning.
\newblock {\em arXiv preprint arXiv:2006.07733}, 2020.

\bibitem{ijcai2020-271}
Yusuke Iwasawa, Kei Akuzawa, and Yutaka Matsuo.
\newblock Stabilizing adversarial invariance induction from divergence
  minimization perspective.
\newblock In Christian Bessiere, editor, {\em Proceedings of the Twenty-Ninth
  International Joint Conference on Artificial Intelligence, {IJCAI-20}}, pages
  1955--1962. International Joint Conferences on Artificial Intelligence
  Organization, 7 2020.
\newblock Main track.

\bibitem{jaiswal2020invariant}
Ayush Jaiswal, Daniel Moyer, Greg Ver~Steeg, Wael AbdAlmageed, and Premkumar
  Natarajan.
\newblock Invariant representations through adversarial forgetting.
\newblock In {\em Proceedings of the AAAI Conference on Artificial
  Intelligence}, volume~34, pages 4272--4279, 2020.

\bibitem{khodabandeh2019robust}
Mehran Khodabandeh, Arash Vahdat, Mani Ranjbar, and William~G Macready.
\newblock A robust learning approach to domain adaptive object detection.
\newblock In {\em Proceedings of the IEEE/CVF International Conference on
  Computer Vision}, pages 480--490, 2019.

\bibitem{kiefer2021leveraging}
Benjamin Kiefer, Martin Messmer, and Andreas Zell.
\newblock Leveraging domain labels for object detection from uavs.
\newblock {\em arXiv preprint arXiv:2101.12677}, 2021.

\bibitem{koga2020method}
Yohei Koga, Hiroyuki Miyazaki, and Ryosuke Shibasaki.
\newblock A method for vehicle detection in high-resolution satellite images
  that uses a region-based object detector and unsupervised domain adaptation.
\newblock {\em Remote Sensing}, 12(3):575, 2020.

\bibitem{lee2019me}
Hyungtae Lee, Sungmin Eum, and Heesung Kwon.
\newblock Me r-cnn: Multi-expert r-cnn for object detection.
\newblock {\em IEEE Transactions on Image Processing}, 29:1030--1044, 2019.

\bibitem{li2014learning}
Yujia Li, Kevin Swersky, and Richard Zemel.
\newblock Learning unbiased features.
\newblock {\em arXiv preprint arXiv:1412.5244}, 2014.

\bibitem{liu2016ssd}
Wei Liu, Dragomir Anguelov, Dumitru Erhan, Christian Szegedy, Scott Reed,
  Cheng-Yang Fu, and Alexander~C Berg.
\newblock Ssd: Single shot multibox detector.
\newblock In {\em European conference on computer vision}, pages 21--37.
  Springer, 2016.

\bibitem{Liu_2018_CVPR}
Yang Liu, Zhaowen Wang, Hailin Jin, and Ian Wassell.
\newblock Multi-task adversarial network for disentangled feature learning.
\newblock In {\em Proceedings of the IEEE Conference on Computer Vision and
  Pattern Recognition (CVPR)}, June 2018.

\bibitem{liu2018multi}
Yang Liu, Zhaowen Wang, Hailin Jin, and Ian Wassell.
\newblock Multi-task adversarial network for disentangled feature learning.
\newblock In {\em Proceedings of the IEEE Conference on Computer Vision and
  Pattern Recognition}, pages 3743--3751, 2018.

\bibitem{lu2019multisource}
Xiaoqiang Lu, Tengfei Gong, and Xiangtao Zheng.
\newblock Multisource compensation network for remote sensing cross-domain
  scene classification.
\newblock {\em IEEE Transactions on Geoscience and Remote Sensing},
  58(4):2504--2515, 2019.

\bibitem{mnih2015human}
Volodymyr Mnih, Koray Kavukcuoglu, David Silver, Andrei~A Rusu, Joel Veness,
  Marc~G Bellemare, Alex Graves, Martin Riedmiller, Andreas~K Fidjeland, Georg
  Ostrovski, et~al.
\newblock Human-level control through deep reinforcement learning.
\newblock {\em nature}, 518(7540):529--533, 2015.

\bibitem{ren2015faster}
Shaoqing Ren, Kaiming He, Ross Girshick, and Jian Sun.
\newblock Faster r-cnn: Towards real-time object detection with region proposal
  networks.
\newblock {\em arXiv preprint arXiv:1506.01497}, 2015.

\bibitem{robins1995catastrophic}
Anthony Robins.
\newblock Catastrophic forgetting, rehearsal and pseudorehearsal.
\newblock {\em Connection Science}, 7(2):123--146, 1995.

\bibitem{rolnick2018experience}
David Rolnick, Arun Ahuja, Jonathan Schwarz, Timothy~P Lillicrap, and Greg
  Wayne.
\newblock Experience replay for continual learning.
\newblock {\em arXiv preprint arXiv:1811.11682}, 2018.

\bibitem{shrivastava2017learning}
Ashish Shrivastava, Tomas Pfister, Oncel Tuzel, Joshua Susskind, Wenda Wang,
  and Russell Webb.
\newblock Learning from simulated and unsupervised images through adversarial
  training.
\newblock In {\em Proceedings of the IEEE conference on computer vision and
  pattern recognition}, pages 2107--2116, 2017.

\bibitem{song2019domain}
Shaoyue Song, Hongkai Yu, Zhenjiang Miao, Qiang Zhang, Yuewei Lin, and Song
  Wang.
\newblock Domain adaptation for convolutional neural networks-based remote
  sensing scene classification.
\newblock {\em IEEE Geoscience and Remote Sensing Letters}, 16(8):1324--1328,
  2019.

\bibitem{tasar2020daugnet}
Onur Tasar, Alain Giros, Yuliya Tarabalka, Pierre Alliez, and S{\'e}bastien
  Clerc.
\newblock Daugnet: Unsupervised, multisource, multitarget, and life-long domain
  adaptation for semantic segmentation of satellite images.
\newblock {\em IEEE Transactions on Geoscience and Remote Sensing}, 2020.

\bibitem{tasar2020standardgan}
Onur Tasar, Yuliya Tarabalka, Alain Giros, Pierre Alliez, and S{\'e}bastien
  Clerc.
\newblock Standardgan: Multi-source domain adaptation for semantic segmentation
  of very high resolution satellite images by data standardization.
\newblock In {\em Proceedings of the IEEE/CVF Conference on Computer Vision and
  Pattern Recognition Workshops}, pages 192--193, 2020.

\bibitem{tenenbaum1997separating}
Joshua~B Tenenbaum and William~T Freeman.
\newblock Separating style and content.
\newblock {\em Advances in neural information processing systems}, pages
  662--668, 1997.

\bibitem{tripathi2019learning}
Shashank Tripathi, Siddhartha Chandra, Amit Agrawal, Ambrish Tyagi, James~M
  Rehg, and Visesh Chari.
\newblock Learning to generate synthetic data via compositing.
\newblock In {\em Proceedings of the IEEE/CVF Conference on Computer Vision and
  Pattern Recognition}, pages 461--470, 2019.

\bibitem{pmlr-v119-wang20h}
Hao Wang, Hao He, and Dina Katabi.
\newblock Continuously indexed domain adaptation.
\newblock In Hal~Daumé III and Aarti Singh, editors, {\em Proceedings of the
  37th International Conference on Machine Learning}, volume 119 of {\em
  Proceedings of Machine Learning Research}, pages 9898--9907. PMLR, 13--18 Jul
  2020.

\bibitem{wang2020few}
Xin Wang, Thomas~E. Huang, Trevor Darrell, Joseph~E Gonzalez, and Fisher Yu.
\newblock Frustratingly simple few-shot object detection.
\newblock In {\em International Conference on Machine Learning}, July 2020.

\bibitem{weir2019spacenet}
Nicholas Weir, David Lindenbaum, Alexei Bastidas, Adam~Van Etten, Sean
  McPherson, Jacob Shermeyer, Varun Kumar, and Hanlin Tang.
\newblock Spacenet mvoi: a multi-view overhead imagery dataset.
\newblock In {\em Proceedings of the IEEE/CVF International Conference on
  Computer Vision}, pages 992--1001, 2019.

\bibitem{wu2019delving}
Zhenyu Wu, Karthik Suresh, Priya Narayanan, Hongyu Xu, Heesung Kwon, and
  Zhangyang Wang.
\newblock Delving into robust object detection from unmanned aerial vehicles: A
  deep nuisance disentanglement approach.
\newblock In {\em Proceedings of the IEEE/CVF International Conference on
  Computer Vision}, pages 1201--1210, 2019.

\bibitem{wu2018towards}
Zhenyu Wu, Zhangyang Wang, Zhaowen Wang, and Hailin Jin.
\newblock Towards privacy-preserving visual recognition via adversarial
  training: A pilot study.
\newblock In {\em Proceedings of the European Conference on Computer Vision
  (ECCV)}, pages 606--624, 2018.

\bibitem{NIPS2017_8cb22bdd}
Qizhe Xie, Zihang Dai, Yulun Du, Eduard Hovy, and Graham Neubig.
\newblock Controllable invariance through adversarial feature learning.
\newblock In I. Guyon, U.~V. Luxburg, S. Bengio, H. Wallach, R. Fergus, S.
  Vishwanathan, and R. Garnett, editors, {\em Advances in Neural Information
  Processing Systems}, volume~30. Curran Associates, Inc., 2017.

\bibitem{xie2017controllable}
Qizhe Xie, Zihang Dai, Yulun Du, Eduard Hovy, and Graham Neubig.
\newblock Controllable invariance through adversarial feature learning.
\newblock {\em arXiv preprint arXiv:1705.11122}, 2017.

\bibitem{xu2020cross}
Minghao Xu, Hang Wang, Bingbing Ni, Qi Tian, and Wenjun Zhang.
\newblock Cross-domain detection via graph-induced prototype alignment.
\newblock In {\em Proceedings of the IEEE/CVF Conference on Computer Vision and
  Pattern Recognition}, pages 12355--12364, 2020.

\bibitem{yazici2019unusual}
Yasin Yazici, Chuan-Sheng Foo, Stefan Winkler, Kim-Hui Yap, Georgios Piliouras,
  and Vijay Chandrasekhar.
\newblock The unusual effectiveness of averaging in gan training, 2019.

\bibitem{yuan2020calibrated}
Ye Yuan, Wuyang Chen, Tianlong Chen, Yang Yang, Zhou Ren, Zhangyang Wang, and
  Gang Hua.
\newblock Calibrated domain-invariant learning for highly generalizable large
  scale re-identification.
\newblock In {\em Proceedings of the IEEE/CVF Winter Conference on Applications
  of Computer Vision}, pages 3589--3598, 2020.

\bibitem{NEURIPS2018_717d8b3d}
Han Zhao, Shanghang Zhang, Guanhang Wu, Jos\'{e} M.~F. Moura, Joao~P Costeira,
  and Geoffrey~J Gordon.
\newblock Adversarial multiple source domain adaptation.
\newblock In S. Bengio, H. Wallach, H. Larochelle, K. Grauman, N. Cesa-Bianchi,
  and R. Garnett, editors, {\em Advances in Neural Information Processing
  Systems}, volume~31. Curran Associates, Inc., 2018.

\bibitem{zhou2019objects}
Xingyi Zhou, Dequan Wang, and Philipp Kr{\"a}henb{\"u}hl.
\newblock Objects as points.
\newblock {\em arXiv preprint arXiv:1904.07850}, 2019.

\bibitem{zhu2020vision}
Pengfei Zhu, Longyin Wen, Dawei Du, Xiao Bian, Qinghua Hu, and Haibin Ling.
\newblock Vision meets drones: Past, present and future.
\newblock {\em arXiv preprint arXiv:2001.06303}, 2020.

\end{thebibliography}
}

\end{document}